\documentclass[sigconf,nonacm]{acmart}
\usepackage{caption}
\usepackage{multirow}
\usepackage{xcolor}
\usepackage{graphicx}
\usepackage{epstopdf}
\usepackage{url}
\usepackage{comment}
\usepackage{graphicx}
\usepackage{caption}
\usepackage{subcaption}
\AtBeginDocument{%
  \providecommand\BibTeX{{%
  \normalfont 
  \kern-0.5em{\scshape i\kern
  -0.25em b}\kern-0.8em\TeX}
  }}
\graphicspath{{plots/}}
\begin{document}

\title{Quantile Regression using Random Forest Proximities}


\begin{comment}
\author{Anonymous Authors}
% \email{Anonymous@}
\affiliation{%
	% \institution{Anonymous}
	% \city{New York}
	% \state{NY}
	\country{Anonymous}
}
\end{comment}

\author{Mingshu Li}
\email{Mingshu.Li@blackrock.com}
\affiliation{%
  \institution{BlackRock, Inc.}
  \city{Atlanta}
  \state{GA}
  \country{USA}
}    
\author{Bhaskarjit Sarmah}
\email{bhaskarjit.sarmah@blackrock.com}
\affiliation{%
  \institution{BlackRock, Inc.}
  \city{Gurgaon}
  \country{India}
}
\author{Dhruv Desai}
\email{dhruv.desai1@blackrock.com}
\affiliation{%
  \institution{BlackRock, Inc.}
 \city{New York}
  \state{NY}
  \country{USA}
}
\author{Joshua Rosaler}
\email{joshua.rosaler@blackrock.com}
\affiliation{%
  \institution{BlackRock, Inc.}
 \city{New York}
  \state{NY}
  \country{USA}
}
\author{Snigdha Bhagat}
\email{snigdha.bhagat@blackrock.com}
\affiliation{%
  \institution{BlackRock, Inc.}
 \city{Gurgaon}
  \state{HR}
  \country{India}
}
\author{Philip Sommer}
\email{philip.sommer@blackrock.com}
\affiliation{%
  \institution{BlackRock, Inc.}
 \city{New York}
  \state{NY}
  \country{USA}
}

\author{Dhagash Mehta}
\email{dhagash.mehta@blackrock.com}
\affiliation{%
  \institution{BlackRock, Inc.}
  \city{New York, NY}
  \country{USA}
}

\renewcommand{\shortauthors}{Li et al.}

\begin{abstract}
Due to the dynamic nature of financial markets, maintaining models that produce precise predictions over time is difficult. Often the goal isn't just point prediction but determining uncertainty. Quantifying uncertainty, especially the aleatoric uncertainty due to the unpredictable nature of market drivers, helps investors understand varying risk levels. Recently, quantile regression forests (QRF) have emerged as a promising solution: Unlike most basic quantile regression methods that need separate models for each quantile, quantile regression forests estimate the entire conditional distribution of the target variable with a single model, while retaining all the salient features of a typical random forest. We introduce a novel approach to compute quantile regressions from random forests that leverages the proximity (i.e., distance metric) learned by the model and infers the conditional distribution of the target variable. We evaluate the proposed methodology using publicly available datasets and then apply it towards the problem of forecasting the average daily volume of corporate bonds. We show that using quantile regression using Random Forest proximities demonstrates superior performance in approximating conditional target distributions and prediction intervals to the original version of QRF. We also demonstrate that the proposed framework is significantly more computationally efficient than traditional approaches to quantile regressions.
\end{abstract}
\begin{comment}
\begin{CCSXML}
<ccs2012>
 <concept>
  <concept_id>10010520.10010553.10010562</concept_id>
  <concept_desc>Computer systems organization~Embedded systems</concept_desc>
  <concept_significance>500</concept_significance>
 </concept>
 <concept>
  <concept_id>10010520.10010575.10010755</concept_id>
  <concept_desc>Computer systems organization~Redundancy</concept_desc>
  <concept_significance>300</concept_significance>
 </concept>
 <concept>
  <concept_id>10010520.10010553.10010554</concept_id>
  <concept_desc>Computer systems organization~Robotics</concept_desc>
  <concept_significance>100</concept_significance>
 </concept>
 <concept>
  <concept_id>10003033.10003083.10003095</concept_id>
  <concept_desc>Networks~Network reliability</concept_desc>
  <concept_significance>100</concept_significance>
 </concept>
</ccs2012>
\end{CCSXML}

\ccsdesc[500]{Computer systems organization~Embedded systems}
\ccsdesc[300]{Computer systems organization~Redundancy}
\ccsdesc{Computer systems organization~Robotics}
\ccsdesc[100]{Networks~Network reliability}

\keywords{Company similarity, Natural Language Processing}
\end{comment}
\keywords{Uncertainty Quantification, Quantile Regression, Random Forest, Proximity}
\maketitle

\graphicspath{ {plots/plots_rf_explainability/} }

\section{Introduction}
Investors make investment decisions based on the potential profit achievable in the future. Numerous studies have focused on forecasting models that forecast a single price for assets with the least error \cite{sezer2020financial,liu2020improved}.  However, the financial market exhibits a high level of volatility and variability, driven by a confluence of economic, political, and technological factors. The time series of trading data is characterized by abrupt fluctuations and volatility, which are often driven by complex internal and external environments. The inherent randomness of these factors often masks underlying trends and leads to poor performance of point estimation models. Moreover, different asset classes (e.g., commodities, bonds, tech stocks) are affected differently by these market forces, exhibiting varying levels of volatility and making it difficult to generalize predictions across asset types (See, e.g., Ref.~\cite{madhavan2022trading} for a recent review). Instead of providing a specific price, often it is more valuable for the investors to have an understanding of the direction and magnitude of potential market movements. By delivering a probable range of future asset prices, investors gain insights into both the trends and the uncertainty associated with these predictions. This range provides crucial information for different investment strategies, whether managing risk-averse portfolios or making speculative bets. It helps quantify the stochastic nature of trading data while accounting for the inherent model bias\cite{kabir2018neural}.In Quantitative liquidity risk management\cite{golub2023blackrock} the key aspect is estimating a bond's "time-to-liquidation" by forecasting average daily volume (ADV) using a model that incorporates historical trading data and various bond features. The modeling process involves estimating both the likelihood of a bond trading and the expected volume if it does, thus calculating the unconditional expected future trade volume.  For frequently traded bonds, historical activity can predict future volume, but for infrequently traded bonds, sparse data may be unrepresentative. In such cases, data from bonds with similar characteristics is used for better predictions. Random forest regression, a machine learning technique, effectively estimates this model by handling nonlinearity, missing data, and naturally incorporating similarities between bonds.

There are typically two types of uncertainties in the context of modeling and predictions: aleatoric and epistemic \cite{der2009aleatory}. Aleatoric uncertainty arises from the inherent randomness within financial data, such as the variability in investor behavior and market prices. This type of uncertainty is irreducible and is a natural aspect of the financial environment. On the other hand, epistemic uncertainty stems from limitations in our knowledge or information regarding the model being used. This could include incomplete historical records or assumptions embedded within the financial models. Quantifying uncertainty, especially the aleatoric component, aims to derive prediction intervals that provide additional insights beyond traditional point estimates. By understanding the extent of risk associated with decision-making, investors can better gauge potential outcomes in volatile markets \cite{huang2018stock, wang2020ensemble}. The literature has seen a rise in studies on the application of uncertainty quantification in finance. The Monte Carlo method, specializes in estimating model outcome uncertainty through repeated random sampling \cite{fan2023uncertainty}. However, this method suffers from high computational costs, particularly when handling high-dimensional, large datasets, limiting its scalability in big data environments\cite{zhang2021modern}. The Bayesian approach is another probabilistic technique used in financial time series forecasting, that is frequently coupled with the Markov Chain Monte Carlo (MCMC) method to update posterior distributions based on prior knowledge\cite{johannes2010mcmc}. Although this approach can incorporate prior information effectively, it is sensitive to model assumptions and highly dependent on the appropriate selection of prior distributions, which can affect the overall performance\cite{wang2020deeppipe}. 

Recently, research has shifted toward more advanced modeling with the rapid advancements in machine learning methods\cite{blasco2024survey, wutte2023take}. Techniques like Bayesian neural networks, autoencoders, and Long Short-Term Memory (LSTM) networks are gaining traction in uncertainty modeling for financial time series, derivative pricing, and portfolio optimization \cite{lim2021time}. Their ability to handle highly nonlinear and complex relationships makes them particularly suited for understanding market dynamics\cite{jang2017empirical}. For instance, Ref.~\cite{chandra2021bayesian} demonstrated the application of Bayesian neural networks for explaining the high volatility in stock price forecasting. Ref.~\cite{xu2018stock} showed the efficacy of recurrent latent variables in managing market stochasticity and proposed a neural network architecture for stock movement prediction. Ref.~\cite{jay2020stochastic} used Multi-Layer Perceptrons (MLP) and LSTM networks to identify market reaction patterns and forecast cryptocurrency prices. Despite their promise, these advanced machine learning models have significant limitations. Their black-box nature and tendency to overfit can make them challenging to interpret and apply reliably in the financial domain\cite{staahl2020evaluation}.

Tree-based methods excel in managing high-dimensional data, resolving imbalanced information, and capturing the dynamics of price trends and volatility. In the literature, several methods have been developed to enhance uncertainty quantification within the random forest framework. One significant approach is the use of bootstrapped confidence intervals within random forests \cite{wager2014confidence, mentch2016quantifying}. By averaging over trees built on subsamples of the training set, the method assesses the variance and confidence levels of the predictions. Additionally, conformal prediction within the random forest framework offers a unique approach. The method calibrates the empirical distribution of the residual on the out-of-bag predictions, and thus providing estimated prediction intervals on the testing data \cite{johansson2014regression}. Furthermore, advancements like inferential framework for Bayesian tree-based regression have been explored. The Bayesian posterior distributions are inferred to construct adaptive confidence bands \cite{castillo2021uncertainty}. Quantile regression forests (QRFs) stand out as an extension of random forests that estimates not only the conditional mean, but the full conditional distribution of the target variable, addressing some limitations of other methods, such as low interpretability, high computational cost, and difficulty handling non-linearity \cite{hastie2009random}.  Introduced by Meinshausen in 2006 \cite{meinshausen2006quantile}, the method operates by aggregating predictions from individual trees within the forest, each tree contributing based on the subset of the training data it has been exposed to. During the prediction phase, a test instance is processed by all trees to identify how much training data from each tree falls into the same terminal nodes. These results are then pooled to form a weighted distribution of the target variable, allowing the conditional quantile to be estimated from this aggregated information.

A recent development in this field is presented in Ref.~\cite{gostkowski2020weighted}, which proposes a weighted approach to enhance traditional quantile regression forests. The authors introduced performance-based weights for quantile estimation, allowing trees that demonstrate superior performance to exert greater influence on the final model, contrasting with the original approach where trees contribute equally regardless of their individual accuracy. Traditional quantile regression methods typically require the training of separate models to estimate each selected quantile of the conditional distribution of the target variable. Although recent advancements such as composite quantile regression \cite{zou2008composite} and multiple quantile modeling via reduced rank regression \cite{lian2019multiple} allow for multiple quantile estimations within a single model, these methods still struggle with modeling the non-linear relationships inherent in the underlying data. QRFs leverage the predictive power of random forests and offer a significant advantage: with just a single model, QRFs can estimate the full conditional distribution of the target variable.  As a result, they avoid the computational cost of training many different models, as well as the possibility that predicted quantiles may not be appropriately ordered on out-of-sample data, which arises in traditional approaches to quantile regression.

Our work introduces a novel instance-based weighting mechanism that leverages random forest proximities. In a nutshell, the proposed method assigns weights at the instance level, based on the similarity between pairs of observations, rather than relying solely on the tree performances. The proximities derived from random forests act as effective local distance metrics on the model’s feature space, providing a robust tool for assessing similarities between observations. These strengths make quantile regression using random forest proximities well-suited to handling noise and variability in data. Integrating these proximities within quantile regression offers a valuable opportunity to enhance predictability while maintaining a balance between accuracy and interpretability. Consequently, we also explore our method using various random forest proximities to improve uncertainty quantification.

Our main contribution in this work are: (1) Develop a novel approach to quantile regression using random forest proximities for uncertainty quantification within financial forecasting tasks; (2) Implement the proposed methodology and benchmark it on publicly available toy datasets; (3) Examine and compare the predictability of quantile regression using different versions of random forest proximities for various public datasets as well as financial applications through a trading volume forecast model.

\section{Quantile Regression using Random Forest Proximities}
In this Section, we describe QRFs, proximities derived from the random forests, and eventually describe our proposed method. 
\subsection{Quantile Regression Forests}
Quantile regression is a traditional tool to estimate the uncertainty, such that the output reflects the upper and lower boundaries of the uncertainty of the unseen data points \cite{hao2007quantile,koenker2017quantile}. While traditionally quantile regressions have been focused on linear models, recently quantile regression methods have been extended to random forests \cite{hastie2009random,meinshausen2006quantile,gostkowski2020weighted}, called quantile random forests (QRFs), that estimate not only the conditional mean $E(y|x)$ of the target variable $y$ given the model input features $x$ (as the conventional random forests do), but the entire conditional probability distribution $p(y|x)$. This density can be estimated by employing numerical weights $\omega_{i}(x)$ associated with each training point (indexed by $i$) that are extracted from the trained random forest model. 

Formally, one can express the cumulative probability $F(y|x)$ associated with the conditional density $p(y|x)$ as the expectation $E(I_{\{Y\leq{y}\}}|X=x)$ of the indicator function $I_{\{Y\leq{y}\}}$, which is equal to 1 for $Y\leq{y}$ and 0 otherwise. The model estimate $\hat{F}(y|X=x)$ for this expectation is computed as a weighted mean over the training observations of $I_{\{Y\leq{y}\}}$:
\begin{subequations}\label{eq:qrf}
\begin{align}
\hat{F}(y|X=x) &= P(Y\leq{y}|X=x) \\
&= E(I_{\{Y\leq{y}\}}|X=x) \\
&= \sum_{i=1}^{n}\omega_{i}(x)I_{\{Y_{i}\leq{y}\}},
\end{align}
\end{subequations}
with 
\begin{subequations}
\label{equations}
\begin{align}
  \label{eq:rfw-2}
  \omega_{i}(x)&=k^{-1}\sum_{t=1}^{k}\omega_{i}(x,\theta_{t}), \\
  \label{eq:rfw-1}
  \omega_{i}(x,\theta)&=\frac{I_{\{X_{i}\in{R_{l}(x,\theta)}\}}}{\#\{j:X_{j}\in{R_{l(x,\theta)}}\}},
\end{align}
\end{subequations}
where $R_{l}(x,\theta)$ is the region of feature space associated with the leaf $l(x,\theta)$ containing point $x$ of the tree parametrized by $\theta$ \cite{meinshausen2006quantile}. In practice, with quantile regression, one estimates quantiles $Q_{\alpha}(x)$, defined by

\begin{equation}\label{eq:quantile}
Q_{\alpha}(x)=inf\{y:F(y|X=x)\geq{\alpha}\}.
\end{equation}

\noindent for some particular values of $\alpha$, where $0 \leq \alpha \leq 1$.

\subsection{Random Forest Proximities}
It was shown in Ref.\cite{lin2006random} that random forests can be mathematically formulated as an adaptive weighted k-nearest-neighbor (kNN) algorithm, in the sense that any prediction of a random forest (which estimates the conditional mean of the target variable), can be expressed as a weighted average of target labels in the training set, with the values of the weights varying across the feature space: 

\begin{equation*}\label{eq:RF_weighted_KNN}
\hat{y}_{i} = \omega_{i}(x) \cdot y_{\mbox{train}}\\
= \omega_{i,1}(x) \ y_{\mbox{train},1} \ + \ ... \ + \ \omega_{i,N}(x) \ y_{\mbox{train},N},
\end{equation*}
where $y_{\mbox{train}, j}$ is the ground truth target label for the $j^{th}$ training example, and $\omega_{i,j}(x)$ is the weight of the observation $j$ in the linear expansion of the RF prediction for observation $i$ with input features $x$. Here, the training points ``nearest" to the test point are understood as those with the largest weights.

\subsubsection{Geometry- and Accuracy-Preserving Proximity}
There are several ways to use random forests to define a notion of ``nearness", ``similarity", or ``proximity" (or inversely, ``distance") between observations in a dataset. All rely in some way on the number of trees in the random forest ensemble for which two points fall in the same leaf node. 

Recently, a definition of proximity based on random forests was proposed that exactly captures the weights $\omega_{i,j}(x)$ in (Eq.~(\ref{eq:RF_weighted_KNN})). This definition takes account of whether a given point in the training set is in-bag or out-of-bag for each tree in the ensemble when attempting to provide an exact closed-form expression for the weights $\omega_{i,j}(x)$. Because these proximities recover the predictions of the random forest exactly, they are also understood to provide a uniquely faithful characterization of the geometry over the feature space encoded in the random forest model, and so are known as Random Forest-Geometry- and Accuracy-Preserving (RF-GAP) proximities \cite{rhodes2023geometry}. 

The GAP proximity, $Prox_{GAP}(i,j)$, is defined as 
\begin{equation}\label{eq:gap}
Prox_{GAP}(i,j) = \frac{1}{|S_{i}|}\sum_{t\in{S_{i}}}\frac{c_{j}(t)\cdot{I(j\in{J_{i}(t)})}}{|M_{i}(t)|},
\end{equation}
where $S_{i}$ is the set of trees where the $i^{th}$ observation is out-of-bag; $c_{j}(t)$ is the multiplicity of the index $j$ in the bootstrap sample associated with tree $t$; $J_{i}(t)$ is the set of in-bag samples that share the same terminal node with the $i^{th}$ observation in the $t^{th}$ tree; $M_{i}(t)$ is the multiset including in-bag repetitions; and $I$ is the indicator function. One can show \cite{rhodes2023geometry} that the proximities in Eq.~(\ref{eq:gap}) are also the unique values for the adaptive kNN weights $\omega_{i,j}(x)$ that exactly recover the predictions of the random forest.

\subsubsection{Other Definitions of Random Forest Proximity}
In the original notion of proximity based on random forests, the proximity score between observations increases by one each time the observations fall into the same terminal leaf node \cite{breiman-cutler-blog}. This count is then normalized by the number of trees $T$ in the forest to yield the final proximity measure (Eq.~(\ref{eq:original})). However, this method does not differentiate between in-bag and out-of-bag data, which may lead to an overestimation of similarity.
\begin{equation}\label{eq:original}
Prox_{Original}(i,j) = \frac{1}{T}\sum_{t=1}^{T}I(j\in{\upsilon_{i}(t)}),
\end{equation}
where $\upsilon_{i}(t)$ contains indices of data records that end up in the same terminal leaf node as $x_{i}$ in the $t^{th}$ tree. 

To address the bias in the original method, the out-of-bag proximity was introduced \cite{hastie2009random,liaw2002classification}. This approach refines the calculation by considering only those data pairs that are out-of-bag samples for each tree, aiming to provide a more accurate representation of the data's underlying structure (Eq.~(\ref{eq:oob})). However,$Prox_{OOB}$ might also introduce bias, as it excludes in-bag observations which are essential for producing random forest predictions.
\begin{equation}\label{eq:oob}
Prox_{OOB}(i,j) = \frac{\sum_{t\in{S_{i}}}I(j\in{O(t)\cap{\upsilon(t)}})}{\sum_{t\in{S_{i}}}I(j\in{O(t)})},
\end{equation}
where $O(t)$ denotes the set of data indices that are out-of-bag samples in the $t^{th}$ tree.

Recently, Random Forest proximities have found various applications in finance \cite{jeyapaulraj2022supervised, desai2023quantifying} as well as in general in case-based explanability of Random Forests \cite{rosaler2023towards}.

\subsection{Quantile Regression using Random Forest Proximities}
For the Random Forests, the proximity matrix quantifies how often pairs of data points end up in the same leaf across different trees in the forest. Data points that more frequently share the same terminal node are more similar to each other, reflecting the learned closeness based on the underlying structure of the data as learned by the forest. Therefore, it can be used to weigh observations for estimating conditional quantiles. When predicting the conditional quantile for a new data point $x_{j}$, the contributions of the training data can be weighted by how close they are to $x_{j}$, as determined by the proximity matrix. This approach ensures that data nearer to $x_{j}$ has more impact on the predicted quantiles.
\begin{equation}\label{eq:qrf}
\hat{F}(y|X=x) = \sum_{i=1}^{n}Prox(j,i)I_{\{Y_{i}\leq{y}\}}.
\end{equation}

\subsection{Evaluation Metrics}
We evaluate our method(s) against the previous methods for QRFs using multiple metrics, including quantile loss, mean squared error (MSE), and mean absolute percentage error (MAPE).
\subsubsection{Quantile Loss}
The quantile loss function evaluates the estimator's performance by the weighted absolute deviations for quantiles from zero to one. A special case of quantile loss is mean absolute error (MAE), when the quantile is equal to 50\%. Let $y$ and $q$ be the observed and predicted values at quantile $\alpha$ of the target variable, then the quantile loss function is defined as in Eq.~(\ref{eq:quantileloss}).
\begin{equation}\label{eq:quantileloss}
L_{\alpha}(y,q) = 
\begin{cases}
  \alpha|y-q| & \text{if $y$ > $q$} \\
  (1-\alpha)|y-q| & \text{otherwise}.
\end{cases} 
\end{equation}
\subsubsection{Mean Squared Error}
The MSE measures the quality of an estimator by the mean of the squared difference between the actual observations $y_{i}$ and predicted values $\hat{y_{i}}$ of the target variable (Eq.~(\ref{eq:mse})).
\begin{equation}\label{eq:mse}
MSE=\frac{1}{n}\sum_{i=1}^{n}(y_{i}-\hat{y_{i}})^2.
\end{equation}
\subsubsection{Mean Absolute Percentage Error}
MAPE is the measure of prediction accuracy obtained by computing the average absolute percentage deviation between the actual observations $y_{i}$ and predicted values $\hat{y_{i}}$ of the target variable (Eq.~(\ref{eq:mape})).
\begin{equation}\label{eq:mape}
MAPE=\frac{1}{n}\sum_{i=1}^{n}\left|\frac{y_{i}-\hat{y_{i}}}{y_{i}}\right|.
\end{equation}

\section{Datasets}
In this Section, we provide details of the public datasets and corporate bonds data utilized in this study.
\subsection{Toy Datasets}
The proposed method was tested on several public datasets for quantile regression tasks to calibrate the implementation and examine the performance. Table \ref{tbl:toy-data} reports a brief summary of the toy datasets used in this work, including Big Mac, Ozone, Boston Housing, etc. 
\begin{table}[ht]
\centering
{\small
\begin{tabular}{cccc}
    \hline
        Data & No. of Instances & No. of Features & Source \\
    \hline
    Big Mac & 69 & 9 & alr3 \\
    Ozone & 203 & 13 & mlbench \\
    Boston Housing & 506 & 14 & mlbench \\
    Abalone & 500 & 9 & UCI\\
    \hline
\end{tabular}
}
\caption{Summary of Public Datasets \label{tbl:toy-data}}
\vspace{-8mm}
\end{table}

\subsection{Corporate Bonds Data}

Corporate bond securities exhibit significant variance in daily observed trade volume, which further translates into considerable uncertainty in estimating time-to-liquidation\cite{golub2023blackrock}. To address this, modeling the distribution of daily traded volume, such as estimating 'Latent Liquidity' at the $90^{th}$ percentile of the ADV distribution, helps quantify this uncertainty in estimating liquidity. The proposed approaches therefore mitigate one of the fundamental problems of measuring asset liquidity as described in Ref.~\cite{golub2023blackrock} by providing a range for time-to-liquidation. Moreover, Regulatory bodies recognize the inherent uncertainties in liquidity analytics and require that managers demonstrate diligent calculation efforts. The proposed methods described here enhance the ability to quantify uncertainties in transaction costs and time-to-liquidation, meeting regulatory expectations and improving analytic precision. While this paper primarily focuses on estimating tradable volume for recently traded bonds, the methodology is adaptable to broader fixed-income trading analytics, including transaction costs and price uncertainty. 

The corporate bonds data is sourced from \textit{MarketAxess TRACE} for U.S. securities. \textit{TRACE} operates under a regulatory reporting system in the U.S. where all eligible trades must be reported and made transparent to the market within 15 minutes. The data used in this study comprises 675,861 records, covering the period from December 1, 2023, to April 4, 2024. The primary objective is to quantify the uncertainty in future trading volumes for corporate bonds.  Daily market-wide bond volumes are highly skewed and exhibit long tails. Thus, the response variable was transformed into logarithmic space to ensure symmetrical error distribution and enable more robust calibration of the random forest.
The inherent unpredictability in bond trading behavior arises from a complex set of factors. Features include:

{\setlength{\parindent}{0cm}\textbf{Amount Out:} Numerical, amount outstanding in USD;}

{\setlength{\parindent}{0cm}\textbf{Max Axe:} Numerical, the highest reported Axe volume, where Axe refers to quotes from dealers for a given security;}

{\setlength{\parindent}{0cm}\textbf{Issuer Amount Out:} Numerical, amount outstanding on issuer level in USD;}

{\setlength{\parindent}{0cm}\textbf{Leh Rising Angel:} Categorical, indicator of a bond whose rating has changed from low grade to high grade;}

{\setlength{\parindent}{0cm}\textbf{Leh Falling Angel:} Categorical, indicator of a bond whose rating has changed from high grade to low grade;}

{\setlength{\parindent}{0cm}\textbf{Years to Maturity:} Numerical, time in years for bond to mature;}

{\setlength{\parindent}{0cm}\textbf{Term:} Numerical, term of the bond in years (e.g. a 5-year bond);}

{\setlength{\parindent}{0cm}\textbf{Amount Issued:} Numerical, amount issued for given bond in USD;}

{\setlength{\parindent}{0cm}\textbf{Age:} Numerical, age in days of the bond;}

{\setlength{\parindent}{0cm}\textbf{Fractional Age :} Numerical, percent completion of bond's lifetime;}

{\setlength{\parindent}{0cm}\textbf{Country US:} Categorical, indicator of US country membership;}

{\setlength{\parindent}{0cm}\textbf{Market US:} Categorical, indicator of US market membership;}

{\setlength{\parindent}{0cm}\textbf{Market Euro:} Categorical, indicator of EURO market membership;}

{\setlength{\parindent}{0cm}\textbf{Market Global:} Categorical, indicator of GLOBAL market membership;}

{\setlength{\parindent}{0cm}\textbf{Leh Aggregation:} Categorical, indicator of membership of a BBG Barc U.S. Aggregate Index;}

{\setlength{\parindent}{0cm}\textbf{Coupon:} Numerical, coupon of the bond;}

{\setlength{\parindent}{0cm}\textbf{Flag Convert:} Categorical, indicator of a convertible period;}

{\setlength{\parindent}{0cm}\textbf{Coupon Type Float:} Categorical, indicator of a floating bond;}

{\setlength{\parindent}{0cm}\textbf{OTR by Maturity:} Categorical, indicator of on-the-run bonds (as measured by maturity);}

{\setlength{\parindent}{0cm}\textbf{OTR by Issuance:} Categorical, indicator of on-the-run bonds (as measured by issue date);}

{\setlength{\parindent}{0cm}\textbf{Index Membership:} Categorical, indicator of membership of a major index;}

{\setlength{\parindent}{0cm}\textbf{Flag 144a:} Categorical, indicator of regulatory 144a status;}

{\setlength{\parindent}{0cm}\textbf{Is Comp Issuer IG:} Categorical.}

\section{Results}
This section presents computational details and results for different QRFs for the toy datasets and the corporate bonds volume forecast problem. 
\subsection{Hyperparameter Optimization and Cross-validation}
The hyperparameters for the quantile regression forests and quantile regression using random forest proximities were optimized using a grid search combined with 5-fold cross-validation. The number of estimators was varied from 50 to 1,000, increasing in varying step sizes. The maximum depth of the trees was explored within a range of 2 to 20. Additionally, the minimum number of samples required at a leaf node and the minimum number of samples needed to split an internal node were searched within the ranges of 2 to 8 and 2 to 10, respectively. The maximum number of features considered for each split was set to the square root of the total number of features. The selected hyperparameters include: 12 for maximum depth, 100 for number of trees, 42 for random seed, and square-root for feature subset strategy. To adapt the corporate bonds data for supervised learning, considering its temporal dependencies, the target variable \textit{TRACE} daily total volume was preprocessed by shifting it backward by one period. A 5-fold sliding window split technique was employed to preserve the temporal order in the data.

\subsection{Quantile Regression using Random Forest Proximities and Prediction Intervals}
The random forest model, optimized with the selected best combination of hyperparameters, was applied to the training splits from 5-fold cross-validation. Proximity measures were extracted and incorporated into Eq.~(\ref{eq:qrf}) to develop quantile predictions across various quantiles for the testing splits. Table \ref{tbl:ql-toy} presents the quantile loss for different methods across these quantiles on toy datasets. Notably, RF-GAP consistently outperformed other methods in terms of quantile loss, demonstrating its superior capability in capturing the underlying geometry and distribution of the data. 

\begin{figure}
\centering
  \includegraphics[width=9cm,height=3.37cm]{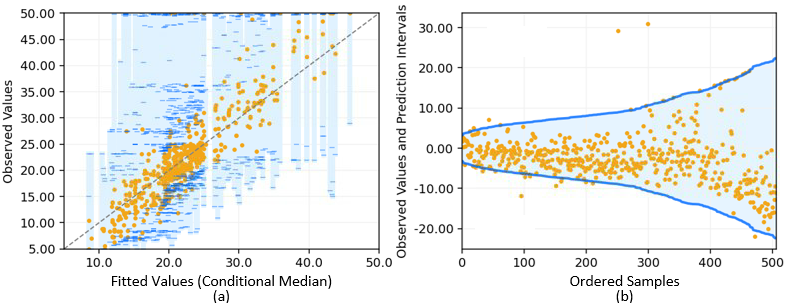}%
  \caption{Forecasting Results of Quantile Regression using RF-GAP (Dataset: Boston). (a): Fitted values based on conditional median, (b): prediction interval at 95\%. }\label{fig: forecast-boston}
\end{figure}

\begin{figure}
\centering
  \includegraphics[width=7.5cm,height=5.32cm]{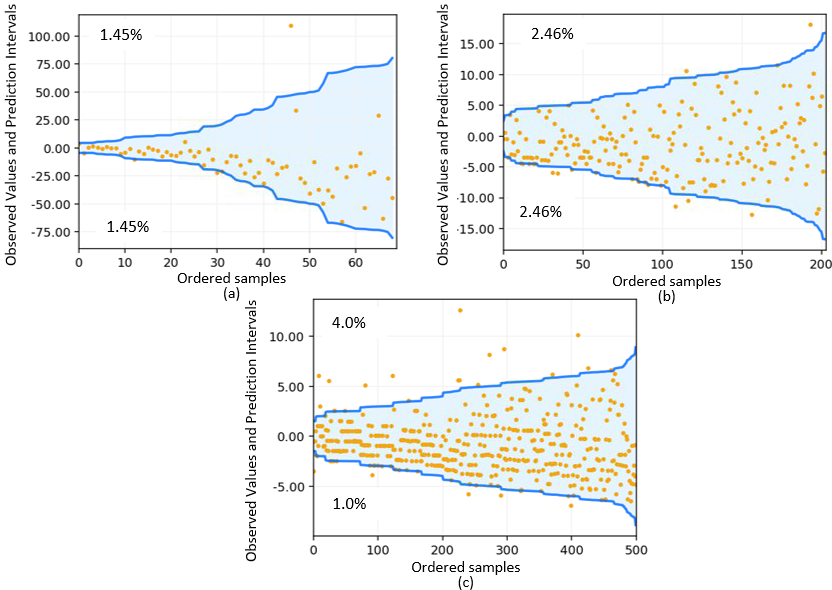}%
  \caption{Prediction Intervals (Quantile Regression using RF-GAP). (a): Big Mac, (b): Ozone, (c): Abalone.}\label{fig: interval-toy}
\end{figure}

The results also allow for confidently predicting the range within which each observation is likely to fall. Figure \ref{fig: forecast-boston} further illustrates these results for one of the toy datasets. On the left, the figure plots the true observations against the predicted conditional median values, with the 95\% prediction intervals shown as transparent blue bars. On the right, the upper and lower bounds of the 95\% prediction interval are plotted against samples ranked by the length of the prediction interval in ascending order, with the mean subtracted for centering around zero. This visualization not only confirms that the prediction intervals approximate the distribution shape of the target variable but also reveals the variability in interval lengths across different data regions. Such variation conveys the confidence associated with each prediction: narrower intervals indicate higher anticipated accuracy compared to wider ones. In addition, Figure \ref{fig: interval-toy} displays plots similar to those on the right side of Figure \ref{fig: forecast-boston} for the remaining datasets. The percentage of data points that fall outside of the prediction intervals is annotated in the upper and lower left corners of each figure. Overall, less than 5\% of the data falls outside these intervals, indicating a high level of accuracy in the model’s predictions across diverse datasets.

\begin{figure}
\centering
  \includegraphics[width=9cm,height=7.24cm]{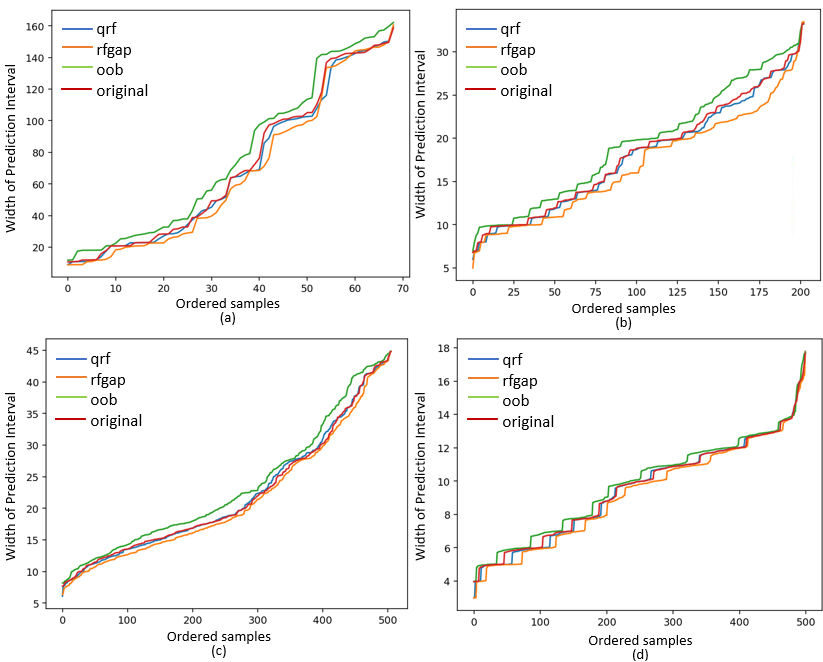}%
  \caption{Width of Prediction Interval (Toy Datasets). (a): Big Mac, (b): Ozone, (c): Boston,(d): Abalone. }\label{fig: interval-length}
\end{figure}

Figure \ref{fig: interval-length} illustrates the lengths of prediction intervals plotted against samples ranked by their interval lengths in ascending order. Across all tested datasets, RF-GAP consistently delivers the tightest prediction intervals compared to other methods, while the Out-Of-Bag (OOB) proximity tends to result in the widest ranges. Table \ref{tbl:ql-toy} complements this analysis by reporting the performance of point estimation based on the conditional median for various methods. Notably, RF-GAP stands out as the top-performing method on the \textit{Ozone} and \textit{Abalone} datasets, achieving the lowest MSE and MAPE. For the \textit{Big Mac} and \textit{Boston datasets}, RF-GAP also secures the lowest MSE scores. Collectively, these outcomes highlight RF-GAP's enhanced proficiency in accurately depicting both the location and dispersion of the conditional distribution of the response variable, markedly outperforming other proximity measures.

Table \ref{tbl:ql-toy} also summarizes the results from the corporate bond volume forecasting model. The performance of the random forest model was evaluated using the median of error metrics across a 5-fold sliding window cross-validation to address the skewed distribution of errors. These results clearly indicate that the RF-GAP method outperforms all other methods across every quantile, showcasing its enhanced capability to capture the variability of trading volumes. Specifically, RF-GAP consistently delivers predictions with lower quantile loss values compared to other approaches, with the original quantile regression forests coming in second, followed by original proximity, and OOB showing the highest quantile loss. This consistent outperformance by RF-GAP reinforces its ability to accurately leverage the underlying data geometry and approximate conditional quantiles more effectively than its counterparts.

\begin{figure}
\centering
  \includegraphics[width=5.5cm,height=4.125cm]{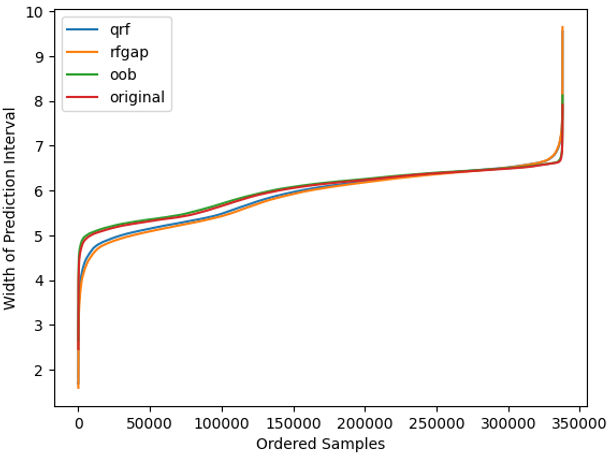}%
  \caption{Width of Prediction Interval (ADV Model)}\label{fig: width-adv}
\end{figure}

\begin{figure}
\centering
  \includegraphics[width=5.5cm,height=4cm]{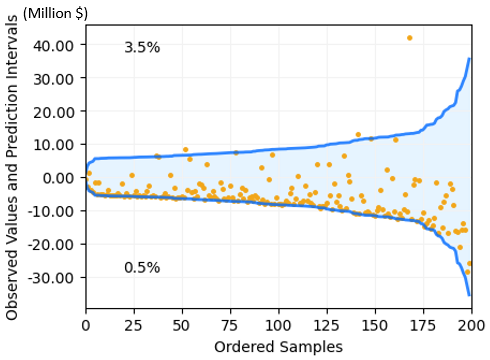}%
  \caption{Prediction Interval of ADV Model (Quantile Regression using RF-GAP)}
  \label{fig: interval-adv}
\end{figure}

Figure \ref{fig: width-adv} examines the width of the prediction intervals across different methods within the logarithmic space. The prediction interval widths range from 2 to 10, indicating a significant variation in uncertainty levels for trading volume forecasts. RF-GAP achieves the narrowest interval widths compared to other methods, pointing to a more precise assessment of predictive uncertainty. This translates to increased confidence and reliability in predictions, which is vital for strategic trading decisions. Lastly, Figure \ref{fig: interval-adv} shows the prediction interval at the 95\% for 200 randomly selected test points using quantile regression using RF-GAP for brevity of presentation. The shape of these intervals, particularly their lower bounds, approximates the true distribution well. The prediction interval spans over 70 million dollars, highlighting substantial uncertainty in future corporate bond trading volumes. This substantial variability indicates the inherent liquidity risk in such investments, highlighting the value of RF-GAP in enabling more informed investment decisions by clearly communicating these risks. 

\begin{table*}[ht]
\centering
{\small
\begin{tabular}{cc|ccccccc|cc}
    \hline
     &  & \multicolumn{7}{c}{Quantile Loss} & \multicolumn{2}{|c}{\begin{tabular}[c]{@{}c@{}}Point Estimation \\ (Conditional Median)\end{tabular}} \\
    \multirow{-2}{*}{Dataset} & \multirow{-2}{*}{Method} & $\alpha$ = 0.005 & $\alpha$ = 0.025 & $\alpha$ = 0.05 & $\alpha$ = 0.5 & $\alpha$ = 0.95 & $\alpha$ = 0.975 & $\alpha$ = 0.995 & MSE & MAPE (\%) \\
    \hline
    \parbox[t]{2mm}{\multirow{4}{*}{\rotatebox[origin=c]{90}{Big Mac}}} &QRF & 0.1306& \textbf{0.5528} & 1.0376 &5.8435&2.7052&2.0824&1.1417 & 502.80 & 25.54\\
    
     &RF-GAP & \textbf{0.1198}& 0.5543 & \textbf{1.0110} &\textbf{5.7442}&\textbf{2.6307}&\textbf{2.0473}&\textbf{1.1265} & \textbf{473.18} & 25.52\\
     
     &OOB & 0.1326& 0.5947 & 1.0847 &5.8584&2.9360&2.2558&1.1760 &524.41&25.64\\
     
     &ORIGINAL & 0.1318& 0.5559 & 1.0466 &5.8501&2.7115&2.1190&1.1486 & 515.54 & \textbf{25.44}\\
     \hline
     
    \parbox[t]{2mm}{\multirow{4}{*}{\rotatebox[origin=c]{90}{Ozone}}} &QRF & 0.0488& 0.2116 & 0.3763 &1.8093&0.4850&0.2867&0.0894 & 20.78 & 48.95\\
    
     &RF-GAP & \textbf{0.0474}& \textbf{0.2075} & \textbf{0.3662} &\textbf{1.7989}&\textbf{0.4844}&\textbf{0.2751}&\textbf{0.0862} & \textbf{20.64} & \textbf{48.68}\\
     
     &OOB & 0.0498& 0.2204 & 0.3898 &1.8619&0.5120&0.3109&0.0942 & 21.78 & 51.30\\
     
     &ORIGINAL & 0.0490& 0.2143 & 0.3774 &1.8174&0.4873&0.2907&0.0902 & 20.87 & 49.43\\
     \hline
     
    \parbox[t]{2mm}{\multirow{4}{*}{\rotatebox[origin=c]{90}{Boston}}} &QRF & 0.0577& 0.2352 & 0.4126 &1.5499&0.6424&0.4321&0.1537 & 22.54 & 15.40\\
    
     &RF-GAP & \textbf{0.0569}& \textbf{0.2317} & \textbf{0.4038} &\textbf{1.5384}&\textbf{0.6300}&\textbf{0.4191}&0.1699 & \textbf{22.13} & 15.38\\
     
     &OOB & 0.0611& 0.2463 & 0.4346 &1.5869&0.6360&0.4503&\textbf{0.1253} & 24.15 & 15.53\\
     
     &ORIGINAL & 0.0587& 0.2378 & 0.4157 &1.5462&0.6327&0.4280&0.1426 & 22.70 & \textbf{15.23}\\
     \hline
     
    \parbox[t]{2mm}{\multirow{4}{*}{\rotatebox[origin=c]{90}{Abalone}}} &QRF & 0.0280& 0.1082 & 0.1972 &0.9573&0.3270&0.2090&0.0781 & 7.39 & 16.34\\
    
     &RF-GAP & \textbf{0.0277}& \textbf{0.1050} & \textbf{0.1948} &\textbf{0.9505}&\textbf{0.3270}&\textbf{0.2085}&0.0790 & \textbf{7.23} & \textbf{16.30}\\
     
     &OOB & 0.0287& 0.1125 & 0.1968 &0.9709&0.3304&0.2124&0.0785 & 7.48 & 16.71\\
     
     &ORIGINAL & 0.0282& 0.1088 & 0.1976 &0.9613&0.3285&0.2092&\textbf{0.0766} & 7.41 & 16.49\\
    \hline

    \parbox[t]{2mm}{\multirow{4}{*}{\rotatebox[origin=c]{90}{\textit{TRACE}}}} &QRF & 0.021044& 0.080092& 0.128156 &0.515995&0.1298&0.070696&0.016540 & 1.0650 & 7.42\\
     
     &RF-GAP & \textbf{0.020902}& \textbf{0.079626} & \textbf{0.127639} &\textbf{0.514303}&\textbf{0.128937}&\textbf{0.070396}&\textbf{0.016471} & \textbf{1.0580} & \textbf{7.40}\\
     
     &OOB & 0.021596& 0.082442 & 0.131357 &0.529950&0.130985&0.070870&0.016604 & 1.1234 & 7.59\\
     
     &ORIGINAL & 0.021526& 0.082142 & 0.130970 &0.528549&0.130332&0.070637&0.016555 & 1.1175 & 7.57\\
    \hline
\end{tabular}
}
\caption{Performance of Quantile Regression using Random Forest Proximities \label{tbl:ql-toy}}
\end{table*}

\begin{table*}[ht]
\centering
{\small
\begin{tabular}{c|ccc|ccc|ccc|ccc}
    \hline
         {\multirow{2}{*}{Method}} & \multicolumn{3}{c|}{$\alpha$ = 0.005} & \multicolumn{3}{c|}{$\alpha$ = 0.025} & \multicolumn{3}{c|}{$\alpha$ = 0.05} & \multicolumn{3}{c}{$\alpha$ = 0.5} \\
         &mse&mae&ql&mse&mae&ql&mse&mae&ql&mse&mae&ql\\
    \hline
    QRF & 0.1280& \textbf{0.1272} & 0.1341 &0.5472&\textbf{0.5440}&0.5877&1.0189&\textbf{1.0057}&1.1039&5.8534&5.8824&\textbf{5.8236}\\
     RF-GAP & \textcolor{red}{0.1190}& 0.1202 & 0.1304 &\textcolor{red}{0.5212}&0.5243&0.5794&0.9883&\textcolor{red}{0.9834}&1.0976&5.7898&5.8372&\textcolor{red}{5.7661}\\
     OOB & 0.1332& \textbf{0.1321} & 0.1385 &0.5831&\textbf{0.5804}&0.6150&1.0793&\textbf{1.1702}&1.1405&\textbf{5.8111}&5.8710&5.8207\\
     ORIGINAL & 0.1290& \textbf{0.1289} & 0.1355 &\textbf{0.5513}&0.5564&0.5933&1.0315&\textbf{1.0275}&1.1172&5.8446&5.8647&\textbf{5.8097}\\
    \hline
    {\multirow{2}{*}{Method}} & \multicolumn{3}{c|}{$\alpha$ = 0.95} & \multicolumn{3}{c|}{$\alpha$ = 0.975} & \multicolumn{3}{c|}{$\alpha$ = 0.995} & \\
    &mse&mae&ql&mse&mae&ql&mse&mae&ql&&&\\
    \hline
    QRF&2.7388&\textbf{2.6620}&2.6750&2.0798&2.0730&\textbf{2.0597}&1.1472&\textbf{1.1468}&1.1563&\\
    RF-GAP&2.8093&2.6778&\textcolor{red}{2.5867}&2.0139&1.9928&\textcolor{red}{1.9913}&\textcolor{red}{1.1174}&1.1262&1.1256&\\
    OOB&3.0061&\textbf{2.9388}&2.9780&2.2596&\textbf{2.2380}&2.2426&\textbf{1.1764}&1.1825&1.2041&\\
    ORIGINAL&2.8175&\textbf{2.7148}&2.7161&2.1216&2.1275&\textbf{2.0976}&\textbf{1.1573}&1.1606&1.1684&\\
    \hline

\end{tabular}
}
\caption{Quantile Loss under Different Random Forest Split Criteria  \label{tbl:custom-criteria}}
\end{table*}

Overall, the results and visualizations confirm that the proposed quantile regression using RF-GAP provides higher accuracy and reliability in uncertainty quantification of future trading volumes. The quantile regression forests employ the same weighting approach as random forests used for approximating conditional mean predictions. Specifically, the learned weights in quantile regression forests are similar to the original proximity method, which treats in-bag and out-of-bag samples with equal weight. In contrast, RF-GAP refines this approach by differentially weighting the in- and out-of-bag samples to account for the random component in the random forest. This method produces an unbiased estimate of weights that more accurately reflects the learned data structure from the random forest and matches the weights used for random forest OOB predictions. By effectively utilizing the data geometry learned from the random forest, quantile regression using RF-GAP delivers robust uncertainty quantification that supports decision-making in volatile markets.

\subsection{Impact of the Split Criterion}
Our experiments were extended to investigate the impact of different split criteria for the random forest algorithm on the performance of conditional quantile estimation. These criteria are used to assess the quality of a split when growing trees in the random forest. Three distinct criteria were tested: mean squared error (MSE), mean absolute error (MAE), and the quantile loss function.

Table \ref{tbl:custom-criteria} provides a summary of the mean quantile loss across 20 random seeds for different split criteria and quantiles. The best-performing results per split criterion are highlighted in bold, while the lowest errors across all methods at each quantile are marked in red. The results indicate that using the quantile loss criterion led to the best performance at the $50^{th}$, $95^{th}$, $97.5^{th}$ quantiles, which in turn yields that a criterion minimizing certain quantile loss does not necessarily lead to reduced quantile loss at that quantile. Moreover, the patterns across different quantiles and datasets did not show consistency. 

\begin{comment}

Figure \ref{fig: custom-criteria} presents the results visually, with error bars illustrating the findings: blue bars represent the MSE outcomes, red bars indicate MAE results, and green bars show the quantile loss criteria results. 

\begin{figure*}
\centering
  \includegraphics[width=16cm,height=9cm]{plots/Error_Bars.png}%
  \caption{Error Bars of Quantile Loss under Different Split Criteria}\label{fig: custom-criteria}
\end{figure*}
\end{comment}
\section{Conclusion}\label{sec:conclusion}
Uncertainty quantification is a vital tool for assessing the volatility and stochasticity inherent in the financial market, enabling investors to better understand the risks associated with their decisions, though a lot of financial forecasting research has focused on point estimation. Despite a growing body of work on uncertainty quantification in finance, many methods still struggle with high computational costs, limited interpretability, and a tendency toward overfitting among other issues. In the present work, a novel quantile regression approach utilizing random forest proximities was proposed, harnessing the strengths of the local distance metrics learned by the random forest to accurately capture the true variability in the response variable. Among the various random forest proximity definitions, the RF-GAP proximity stands out for its ability to preserve the learned geometry of the random forest and effectively measure data distances\cite{rhodes2023geometry}. The proposed methods were tested on a range of public datasets and corporate bond data. Compared to the original quantile regression forests and other proximity measures, the quantile regression using RF-GAP consistently delivered the lowest quantile loss and the narrowest prediction intervals across multiple datasets and quantiles. This more compact prediction interval indicates higher confidence in the model's point estimates, while superior performance in terms of MAE, MSE and quantile loss indicates greater reliability of RF-GAP-based quantile regression in capturing the true variability and conditional quantiles of the response variable.

This research advances the existing body of knowledge by: a) developing a novel quantile regression approach using random forest proximities to improve uncertainty quantification; and b) rigorously evaluating and benchmarking its performance on various public datasets and corporate bond data. The quantile regression approach with random forest proximities, particularly RF-GAP, is expected to offer invaluable insights for decision-makers, helping them gain a deeper understanding of embedded uncertainty and risk. This enhanced understanding is crucial for developing more informed investment strategies that account for market volatility and the associated levels of risk. By accurately estimating prediction intervals, decision-makers can better anticipate and mitigate potential market shifts. Out of various financial applications that would benefit from this, this paper chose liquidity risk to exemplify the principals that would equally work to determine price uncertainty. Furthermore, the proposed method can be applied to other financial applications, such as portfolio optimization, option pricing, and credit risk assessment, to validate the model's robustness and generalizability across different domains.

We also plan to extend the similarity-based framework to conformal predictions and other uncertainty quantification methods.

\section{Acknowledgement}
The views expressed here are those of the authors alone and not of BlackRock, Inc. We thank Donald Edgar and Pratik Aswani for their valuable feedback to this work.

%BlackRock, Inc.
\bibliographystyle{unsrt}

\bibliography{sample-base}
\end{document}